\documentclass{article} 
\usepackage{proceedings}
\usepackage{latexsym}
\usepackage{amsfonts}
\usepackage{named}

\newtheorem{theorem}{Theorem}
\newtheorem{ex}{Example}
\newenvironment{example}{\begin{ex} \rm}{$\Box$ \end{ex}}

\newcommand{\default}[3]{\frac{#1:#2}{#3}}

\newcommand{\lc}{\ulcorner}
\newcommand{\rc}{\urcorner}

\title{Evaluating Defaults}
\author{
{\bf Henry E. Kyburg, Jr.}$^{1,2}$\\
{\tt kyburg@cs.rochester.edu}
\vspace{1ex}\\
$^{1}$Computer Science and Philosophy\\
University of Rochester\\
Rochester NY 14627, USA
\And
{\bf Choh Man Teng}$^{2}$\\
{\tt cmteng@ai.uwf.edu}
\vspace{1ex}\\
$^{2}$Institute for Human and Machine Cognition\\
University of West Florida\\
Pensacola FL 32501, USA
}

\begin{document}

\maketitle

\begin{abstract}
We seek to find normative criteria of adequacy for nonmonotonic logic similar
to the criterion of validity for deductive logic.  Rather than stipulating that
the conclusion of an inference be true in all models in which the premises are
true, we require that the conclusion of a nonmonotonic inference be true in
``almost all'' models of a certain sort in which the premises are true.  This
``certain sort'' specification picks out the models that are relevant to the
inference, taking into account factors such as specificity and vagueness,
and previous inferences.  The frequencies
characterizing the relevant models reflect known frequencies in our actual
world.  The criteria of adequacy for a default
inference can be extended by thresholding to criteria of adequacy for an
extension.  We show that this avoids the implausibilities that might otherwise
result from the chaining of default inferences.  The model proportions,
when construed in terms of frequencies, provide a verifiable grounding of
default rules, and can become the basis for generating default rules from
statistics.

\end{abstract}

\section{Introduction}

Non-monotonic reasoning, for example default logic~\cite{Reiter80},
models the intuitive process of making non-deductive
inferences in the face of certain supportive but not conclusive evidence.
Given a default theory $\Delta=\langle D,F \rangle$, we can obtain its
extensions by following a prescribed set of steps.
However, on what grounds do we employ a particular default rule?
Some writers would
regard this as an inappropriate question, since they take as their goal the
representation of human inference.
To this end defaults represent rules that we take to be intuitively
appropriate.  But then when we apply these rules,
we may be led to counterintuitive
results~\cite{ReiterC81,Lukaszewicz88,Poole89}.
The underlying principle seems to be circular: the original default
rules are ``intuitively good'' at first glance, but when we discover that they
do not give rise to the desired results, we tweak the rules until they give us
those results. It seems that we have to know what results we want first before
constructing the default theory, rather than having the default theory tell us
what conclusions are warranted. This is precisely the reason why we need an
independent measure of validity for default rules and default extensions.
We think of nonmonotonic
logic as sharing the normative character of other logics.  From this point of
view default rules require some defense.
We will concentrate on default logic here, though much of
what we have to say will apply to other nonmonotonic approaches as well.
Much
of the work on nonmonotonic logic has concerned the syntactic manipulation
of the nonmonotonic rules,
rather than their basic justification.

\subsection{Selective Preference}
For a {\em default rule\/}
$d=\default{\alpha}{\beta_1, \ldots, \beta_n}{\gamma}$,
$\alpha$ is the {\em prerequisite\/},
$\beta_1, \ldots, \beta_n$ are the {\em justifications\/},
and $\gamma$ is the {\em consequent\/} of $d$.
Loosely speaking, the rule
conveys the idea that if $\alpha$ is provable, and
$\neg\beta_1, \ldots, \neg\beta_n$ are
each
not provable,
then  by default  we conclude that $\gamma$ is true.
A {\em default theory\/} is an ordered pair
$\langle D,F \rangle$, where
$D$ is a set of default rules and $F$ is a set of
``facts''.
A theory extended from $F$ by applying the default rules in $D$
is known as an {\em extension\/} of the default theory.

Consider the following canonical example.

\begin{example}
We have a default theory $\Delta=\langle D, F \rangle$, where
\[\begin{array}{rcl}
D &=& \{\default{R(x)}{T(x)}{T(x)},
     \default{S(x)}{\neg T(x)}{\neg T(x)}\}, \\
F &=& \{R(a),S(a)\}.
\end{array}\]

We get two extensions, one containing $T(a)$ and the other containing
$\neg T(a)$.
If we take ``$R(x)$'' to mean that $x$ is a bird, ``$S(x)$'' to mean that
$x$ is a penguin, and ``$T(x)$'' to mean that $x$ flies, then we would like
to reject the extension containing ``$T(a)$'' ($a$ flies) in favor of the
extension containing ``$\neg T(a)$'' ($a$ does not fly).
However, if we take ``$S(x)$'' to mean that $x$ is an animal instead and
keep ``$R(x)$'' and ``$T(x)$'' the same, we would want to reverse our preference.
Now the extension containing ``$T(a)$'' ($a$ flies) seems better.
\end{example}
Note that each of the default rules involved in the example above
is intuitively appealing
when viewed by itself against our background knowledge:
 birds fly;
penguins do not fly;
and animals in general do not fly either.
Moreover, both instantiations (penguins and animals) are syntactically
identical.  Thus, we cannot base our decision to prefer one default rule over
the other by simply looking at their syntactic structures.

It is the interaction between the default rules and evidence
that gives rise to the selective preference above.  We have
the evidence that ``$a$ is a bird''.
If in addition we also have ``$a$ is a penguin'',
we prefer the penguin rule.
If instead we have ``$a$ is an animal'', we prefer the bird rule.

There are several approaches to circumventing this conceptual difficulty.
The first is to revise the default theory so that the desired result
is achieved~\cite{ReiterC81}.  We can amend the default rules
by adding the exceptions as justifications, for example
$\default{B(x)}{F(x), \neg P(x)}{F(x)}$ and
$\default{A(x)}{\neg F(x), \neg B(x)}{\neg F(x)}$.
With this approach we have to constantly revise the default rules
to take into account additional exceptions.
We have little guidance in constructing the list of justifications
except that the resulting default rule has to produce the ``right''
answer in the given situation.

Another approach is to establish some priority structure over the
set of defaults.
For example, we can refer to a specificity or inheritance hierarchy
to determine which default rule should be used in case of a
conflict~\cite{Touretzky84,HortyTT87}.
The penguin rule is more specific than the bird rule, when both are
applicable, and therefore we use the penguin rule and not the bird rule.
However, conflicting rules do not always fit into neat hierarchies
(for example, adults are employed, students are not, how about adult
students?~\cite{ReiterC81}).  It is not obvious how we can extend
the hierarchical structure without resorting to explicitly enumerating
the priority relations between the default rules~\cite{Brewka89,Brewka94}.

The third approach is to appeal to probabilistic
analysis.  Defaults are interpreted as
 representing  properties of conditional probabilities.
For example, the conditional probability of $a$ being able to fly given
that $a$ is a bird is ``high''~\cite{Pearl88,Pearl90}
or increases from the prior probability~\cite{NeufeldPA90},
while the conditional probability of $a$ being able to fly given that $a$ is 
a penguin
is ``low'' or decreases.  This approach provides a probabilistic semantics
for default rules, but in a way which does not represent the fact that
 the conclusions are {\em accepted\/}.  The default conclusion is ``Tweety 
flies,'' not ``Probably Tweety flies.''
  This is in contrast to the spirit of
nonmonotonic reasoning: default conclusions should be accepted as new facts,
and we should be able to chain default rules
and build upon the conclusions of previous default applications
to obtain further conclusions.

\subsection{Justifying Nonmonotonic Inference}

The justification of beliefs is a long standing issue in epistemology.  There
is
not much that is problematic about the justification of beliefs obtained by
deductive inference (though there are plenty of problems that surround
deduction --- see
\cite{kyburg.justification,haack.justification,dummett.justification}, not to
mention the voluminous literature on paraconsistent logic \cite{priest}).  The
reason is that we can show that the ordinary rules of deduction lead from
premises to conclusions that are true in every model in which the premises are
true.  This is exactly what is not true of ampliative inference, and it is what
has led some writers (e.g., \cite{morgan}) to deny that there {\em is} any such
thing as a nonmonotonic logic.  This has been disputed in \cite{kyburg.2001}.

But other kinds of justifications of beliefs have been proposed.  Isaac Levi
\cite{levi.gambling,levi.enterprise,levi.argument} has argued for many years
that the way to understand ampliative (inductive, nonmonotonic) argument is in
terms of decision theory: we choose (decide) to accept a hypothesis in a given
context provided that the expected epistemic utility of doing so in that
context
is greater than the expected utility of any other epistemic act, such as
suspending belief totally, or accepting a stronger hypothesis.

Levi's approach employs a rich and detailed structure for acceptance, and
allows drawing many important distinctions.    This structure requires three
things that make it less than perfect as a vehicle for ordinary nonmonotonic
inferential systems.  First, in keeping with a long tradition in pragmatism
\cite{dewey,peirce,james} the context of inquiry must be tied to a specific
problem: We need the answer to a question.  Second, the epistemic expectation of an
answer is the expected value of the {\em information} contained in that
answer.  Thus we need to presuppose an information measure on the language of
our inquiry \cite[p. 169]{levi.argument}. Third, we need to have available a
credal or inductive probability,  based on a measure (or
convex set of measures) on the sentences of the language, in terms of which a
conditional probability (or convex set of conditional probabilities)
can be defined  \cite[p. 52]{levi.enterprise}.

It is our belief 
both that in some contexts in which we might wish to use nonmonotonic
mechanisms, this overhead is unnecessary, and perhaps itself difficult to
justify, and that we would like to be able to explicate the justification of
inference in a less context dependent way.

Another approach that has attracted considerable
attention in the philosophical community in recent years is that of
``reliabilism'' whose best known exponent is Alvin Goldman \cite{goldman}.
According to this view, what justifies a belief is the fact that it is obtained
by a ``reliable cognitive process...'' \cite[p. 20]{goldman}  Of course there
are a number of additional hedges to the view that are required for philosophical
accuracy, and even with those hedges there remains a certain vagueness in the
view.  These details need not detain us, since we are seeking inspiration
rather than philosophical precision.

What does ``reliable'' mean?  
We will construe reliability in terms of frequency or propensity to yield truth
 when applied.
Specifically, we will say that the belief $\phi$ is nonmonotonically justified
by a default rule if the rule would frequently lead to truth and
rarely to error, given what we know --- given our background knowledge.

A deductive argument is justified (valid) if its conclusion is true in every
model of its premises.
We will attempt to provide an analog of the justification of deductive rules:
a default argument is justified if its
conclusion is true in a high proportion of the relevant models in which its premises are
true.
To make this idea precise requires an excursion into model theory.

\section{Model Theory}

We will suppose that the underlying object language is a first order
 language that does not involve such intensional predicates
as ``know'' or ``believe.''  A number of nonmonotonic formalisms
(specifically autoepistemic logic~\cite{Moore85})
do involve such locutions within the object
language, but they can be dispensed with
 in default logic.  The default rule
$\default{\alpha}{\beta_1,\ldots, \beta_n}{\gamma}$ can be read in terms
of the nonmembership of $\ulcorner \neg \beta_i \urcorner$ in a specified set
of
expressions $\Gamma$. In original default logic, $\Gamma$ would
just be an extension.

 There are a number of immediate problems associated
with the idea of looking at the ``proportion'' of models.
The least of them is choosing a {\em level\/} at
which to regard the evidence as adequate.  Should we require that the
proportion be 0.95?  Or 0.99?  Or 0.995?  This is
just the sort of question that
arises in statistical hypothesis testing or in confidence interval estimation.
We shall suppose that in a given context there is some
agreed-upon level of security $\delta$; we will accept a conclusion if the
proportion of models in which we could be committing an error is no greater
than $\delta$.

This approach is to be contrasted with those of Adams \cite{adams,adams.1966},
Pearl \cite{Pearl88} and Bacchus et al \cite{BacchusGHK93}.  Adams requires that
for
$A$ to be a reasonable consequence of the set of sentences $S$, for {\em any\/}
$\epsilon$ there must be a positive $\delta$ such that for every probability
function, if the probability of every sentence in $S$ is greater than $1 -
\delta$, then the probability of $A$ is at least $1 - \epsilon$ \cite[p.
274]{adams.1966}.  Pearl's approach similarly involves quantification over
possible probability functions.  
Bacchus et al again take the degree of belief of a statement to be
the limiting proportion of first order models in which the statement is true.
All of these approaches involve matters that go well beyond what we may
reasonably suppose to be available to us as empirical enquirers.  Our $\delta$, on
the other hand, serves much like the $\alpha$ of statistical testing.

We must restrict the number of models
under consideration to a finite  number so that the idea
of looking at proportions makes
sense.\footnote{We could, instead, seek to develop a
way of proceeding to a limit; this still would require restrictions to arrive
at a countable number of models, and would entail a large expository cost for
little gain in plausibility.}
We will be taking account of statistical information, and to this end will want
each model to have a finite domain.
  Roughly speaking, we take as a model
of our language one in which the domain of empirical individuals is of finite
cardinality.  This may be regarded as problematic (it
entails the falsity of ``every person has two parents  and
nobody is his own ancestor'') but with reasonable spatial and temporal bounding
it can be rendered plausible.

Even so, to ensure that the set of models is finite
we must restrict the empirical domain even further.  Not only must it consist
of
a finite set of physical entities, but this same set of physical entities
$\mathcal D$ must be taken to be the empirical domain of every model.

We assume that it is possible to express statistical knowledge in this
language.  For example, if ``$B(x)$'' is the predicate ``is a bird''
and
``$F(x)$'' is ``can fly'', we can express the fact that
between 85\% and 95\% of birds fly by the formula $0.85 <
\frac{|\{x:B(x)
\land F(x)\}|}{|\{x:B(x)\}|} < 0.95$.  Employing the notation of
\cite{kyburg.teng} we write this as ``$\%x(F(x),B(x),0.85,0.95)$.''
This renders ``\%''  a variable binding operator on 4-sequences of expressions: 
two formulas and two fractions.

We distinguish, as do Pearl and Geffner \cite[p. 70]{pearl.geffner} between
immediate evidence, represented by a finite set of sentences $E$ concerning
particular facts (to be distinguished  from the general body of factual
knowledge $F$ invoked by classical default logic), and a finitely axiomatizable
set of sentences
$K$ representing general background knowledge.
 What defaults are plausible depends, of course, on background knowledge.
If it were not for what we take to be the typical (or natural, or frequent)
behavior of birds, the world's best known example of a default rule would not
be
plausible.  On the other hand no one has proposed the default rule $\frac
{\mathrm{fish}(a):\mathrm{mackerel}(a)}{\mathrm{can-talk}( a)}$.

Thus in general
we will represent the set of default rules of a default theory as $\Delta_K$
rather than $D$, since we take them to be a function of our body of general
knowledge $K$. Given an error tolerance 
$\delta$, we will take a default rule to be {\em $\delta$-valid} if, for every
set of possible input sentences $E$ consistent with $K$, the application of the
rule to $E$ leads to a false conclusion in a
proportion of at most
$\delta$ of the relevant models.  
More precisely, a default rule is $\delta$-valid if and only if for every set of
input sentences $E$ consistent with $K$ to which the rule is applicable, the
proportion of models of $E \cup K$ in which the conclusion of the rule is false
is no more than $\delta$.

To fix our ideas, let us begin with a simple example.  Suppose
$K$ includes a statement to the effect that at least $1 -
\delta$ and not more than 1 - $\epsilon$ of birds fly and nothing else; that is,
``$\%x(F(x),B(x),1-\delta,1-\epsilon)$.''  Consider the rule
$\default{B(x)}{F(x)}{F(x)}$.
This
rule is ``applicable'' to immediate evidence 
$E$ only if $E \cup K$ entails a sentence of the form
$\ulcorner B(a) \urcorner$ and no corresponding sentence of the form
$\ulcorner \neg F(a) \urcorner$.

Our models have a single domain $\mathcal D$ of finite cardinality.
We will write ``$\mathcal I_m(\phi)$'' for the interpretation of $\phi$ in
the model $m$.  The constraint imposed by
$K$ is that for every model $m$ the proportion of objects in $\mathcal I_m(B)$ that
are
also in $\mathcal I_m(F)$ lies in
$[1-\delta,1-\epsilon]$.

There are three cases.  First, suppose $E \cup K$ does not entail a sentence of
the form $\ulcorner B(a) \urcorner$.  Then the rule is inapplicable.  Second, 
suppose that for some term $a$, $E \cup K$ entails
``$B(a)$'' and also entails ``$\neg F(a)$''.  The rule is again inapplicable, because
it is blocked by the failure of a justification.  Third, suppose
for some term $a$, $E \cup K$ entails
``$B(a)$'' but not ``$\neg F(a)$''.
Then $\mathcal I_m(a) \in \mathcal I_m(B)$.  There are
$|\mathcal I_m(B)|$ interpretations of $a$ that make $E \cup K$ true;
of these at least
$1-\delta$ make ``$F(a)$'' true.  We have said nothing about interpreting the
rest of the language, but however many interpretations there are (we have seen
to it that there are only a finite number) the proportion that renders
``$F(a)$'' true will remain unchanged; it will be at least $1-\delta$. 
 Thus, given the background knowledge that we have
posited, the rule is $\delta$-valid: if it is applicable it will lead to error no 
more than $1- \delta$ of the time.

Now let us consider a somewhat more complex example:  Suppose we know
that typically birds fly, and that typically penguins don't.  If that is in our
background knowledge $K$, as well as ``$\forall x(P(x) \supset B(x))$'', then
the
flying default becomes $\default{B(x)}{F(x),\neg P(x)}{F(x)}$, and
we also have the default
$\default{P(x)}{\neg F(x)}{\neg F(x)}.$
If $E$ entails ``$P(a)$'', only the second default is applicable.  In no more
than
$\delta$ of the models of $E$ will $a$ fly, unless $E \cup K$ entails that $a$ can
fly.

Another example: Suppose $K$ contains vague information about the
frequency with which red birds fly (perhaps because we have 
encountered few red birds).  
Say that we know the frequency to be between
0.50 and 1.0. 
Since the interval for birds in general
$[1-\delta,1-\epsilon]$ is included in $[0.5, 1.0]$,
this additional piece of information
should not interfere with our inference of flying ability.
There is no {\em conflict\/} between the two intervals,
just less precision in one.  The rule about birds in general
can be applied to red birds.  However, if $K$ contains the knowledge
that
between 0.5 and
$r$ of red birds fly, where
$r$ is less than $1-\epsilon$, 
then this information
{\em should\/} interfere.  
In this case the general rule should be so construed that it does not 
apply to red birds.  If $b$ is a red bird and not a penguin, no conclusion about
 flying ability is justified.

We can arrange this by judiciously adding or deleting
justifications in the general bird rule, in accordance with the statistical
information in $K$:
in the first case we allow red birds; in the second we must require that we do
not know $a$ is red: ``$\neg R(x)$'' must be among the justifications of
the rule.  This statistical approach provides exactly the normative guidance
that is lacking in the ad-hoc
approach of tweaking default rules in order to arrive at the ``intuitive''
results.

More generally, we can give recipes for constructing $\delta$-valid defaults
for conclusions of the form $\lc \phi (a) \rc$ from background knowledge $K$ 
and immediate evidence $E$.\footnote{
Of course any default conclusion can be given this form, particularly if we allow
the term $a$ to be an $n$-sequence of terms.  Furthermore, any such conclusion
can be taken to be an instance of the consequent of a statistical generalization,
in virtue of the fact that statistical generalizations merely impose bounds. We
are  not imposing serious limitations on default rules.  For details, see
\cite{kyburg.teng}.}
Let $K$ contain $\ulcorner
\%x(\phi(x),\psi(x),p,q) \urcorner$ and 
$\ulcorner
\%x(\phi(x),\psi'(x),p',q') \urcorner$.
We consider three cases:
\begin{enumerate}
\item  $K$ entails $\ulcorner\forall x
(\psi(x) \supset \psi'(x)) \urcorner$.
There are three subcases
according to the relation among $p$, $p'$, $q$, $q'$:
\begin{enumerate}
\item ($p \leq p'$ and $q \leq q'$) or ($p' \leq p$ and $q' \leq q$)\\
$\default{\psi(x)}{\phi(x)}{\phi(x)}$ and $\default{\psi'(x)}{\neg
\psi(x),\phi(x)}{\phi(x)}$ are candidate defaults.
\item $p \leq p'$ and $q' \leq q$\\
$\default{\psi'(x)}{\phi(x)}{\phi(x)}$ is the only candidate default,
since
the justification $\neg \psi'(x)$ 
of $\default{\psi(x)}{\phi(x),\neg \psi'(x)}{\phi(x)}$ is
inconsistent with the prerequisite $\psi(x)$ and $K$.
\item $p' \leq p$ and $q \leq q'$\\
$\default{\psi(x)}{\phi(x)}{\phi(x)}$ and $\default{\psi'(x)}{\neg
\psi(x),\phi(x)}{\phi(x)}$ are candidate defaults.
\end{enumerate}
\item $K$ entails $\ulcorner\forall x
(\psi'(x) \supset \psi(x)) \urcorner$.
This is symmetrical to case 1.
\item $K$ entails neither $\ulcorner\forall x
(\psi(x) \supset \psi'(x)) \urcorner$ nor  $\ulcorner\forall x
(\psi'(x) \supset \psi(x)) \urcorner$.
Again there are three subcases:
\begin{enumerate}
\item ($p \leq p'$ and $q \leq q'$) or ($p' \leq p$ and $q' \leq q$) \\
The candidate defaults are $\default{\psi(x)}{\phi(x),\neg \psi'(x)}{\phi(x)}$
and
$\default{\psi'(x)}{\neg \psi(x),\phi(x)}{\phi(x)}$.
\item $p \leq p'$ and $q' \leq q$\\
$\default{\psi(x)}{\phi(x),\neg \psi'(x)}{\phi(x)}$ and
$\default{\psi'(x)}{\phi(x)}{\phi(x)}$ are candidate default rules.
\item $p' \leq p$ and $q \leq q'$:
This is symmetrical to case 3(b).
\end{enumerate}
\end{enumerate}

Having generated a list of candidate default rules based on our background
knowledge $K$, we delete those rules derived from statistics with a lower
measure less than $1 -\delta$.  The remainder is the set of
defaults $\Delta_K$.

We have not taken account of relations among default conclusions that may be
entailed by $K$.  If $K$ contains $\ulcorner \forall x
(\phi(x) \equiv \phi'(x)) \urcorner$ then the default conclusion $\ulcorner
\phi(a) \rc$ behaves just like the default conclusion $\lc \phi'(a) \rc$.  If
$K$ contains $\lc \forall x(\phi(x) \supset \phi'(x)) \rc$, then since $\lc
\phi(x) \rc$ is equivalent to $\lc \phi(x) \wedge \phi'(x) \rc$ and $\lc
\phi'(x) \rc$ is equivalent to $\lc \phi(x) \vee \phi'(x) \rc$ we can make use
of the obvious entailment relations.

Soundness of a system of deductive logic requires that the conclusion of any 
inference be true in every model in which the premises are true.  Clearly
nonmonotonic inference should not be sound.  But there is a property that is
{\em like\/} soundness that applies to default inference.  It is the property
that the conclusion is false in at most a fraction $\delta$ of the models of the
premises $K \cup E$.

\begin{theorem}[Default Soundness]
For every set of observations $E$, if $d \in \Delta_K$ is applicable to $E$,
the proportion of models of $E \cup K$ in which the conclusion of $d$ is false
is less than $\delta$.

{\rm The proof of this theorem is provided by the soundness theorem for
evidential probability \cite[p. 241]{kyburg.teng}, since the rules for deriving
defaults are a subset of the rules for computing evidential probabilities.
$\Box$}
\end{theorem}

\section{Interactions within an Extension}

Having determined which default rules are justified with respect to the
background knowledge, the next step is to investigate the interaction between
default rules in generating an extension.
A default extension is a minimal deductively closed set that
contains the given facts and the consequents of all applicable default rules.
Given an evidence set, we need to determine
how to control the compound effects of multiple defaults in an extension.

Take for example, a default version~\cite{Poole89} of
the probabilistic lottery paradox~\cite{Kyburg61}.  There are $n$ species
of birds, $S_1, \ldots, S_n$.
We can say that penguins are atypical in that they cannot fly;
hummingbirds are atypical in that they have very fine motor control;
parrots are atypical in that they could talk; and so on.
If we apply this train of thought to all $n$ species of birds,
there is no typical bird left, as for each species
there is always at least one aspect in which it is atypical.
A parallel scenario is formulated below.

\begin{example}
\label{ex:birds}
$K$ contains
\begin{quote}
$B(x) \equiv S_1(x) \vee \ldots \vee S_n(x)$\\
\mbox{\ } \hfill
[an exhaustive list of bird species]\\
$S_i(x) \supset \neg S_j(x)$, for all $j \neq i$\\
\mbox{\ } \hfill
[species are mutually exclusive]\\
$\%(S_i(x), B(x), \epsilon_i, \delta_i), \mbox{for } 1 \leq i \leq n$\\
\mbox{\ } \hfill
[the proportion of each $S_i$ species of birds is ``small'']
\end{quote}
From $K$ we can derive $n$ $\delta^*$-valid default rules for $\Delta_K$:
\[d_i=\default{B(x)}{\neg S_i(x)}{\neg S_i(x)}, \mbox{for } 1 \leq i \leq n\]
where $\delta^*$ is  the maximum of $\delta_1, \ldots, \delta_n$.

Now consider the evidence set $E=\{B(a)\}$.
In the original formulation of default logic,
we would get $n$ extensions, each one containing one $S_i(a)$ and the
negations of all the other $S_j(a)$'s.
Thus, for each extension, we would conclude that $a$ is a particular species
of bird, which seems to be an over commitment, considering we have
$\%(S_i(x), B(x), \epsilon_i, \delta_i)$ in $K$.
\end{example}
Note that each of the $n$ default rules is $\delta$-valid when considered
individually, but in an extension the rules
interact to sanction a set of conclusions that when taken together
seems implausible according to our knowledge of model frequencies.
The definition of an extension dictates that we must keep applying
rules until all ``applicable'' ones are exhausted.  The
``applicability'' condition is based on maximizing logical strength:
for $d=\default{\alpha}{\beta_1, \ldots, \beta_m}{\gamma}$,
as long as $\alpha$ is derivable, and the $\beta$'s are consistent with
the extension, we must apply $d$ and add $\gamma$ to the extension.
Thus, for each of the extensions above, we have to keep applying the rules
until we have drawn $n-1$ conclusions: $\neg S_j(a)$ for all $j \neq i$.
Then the consistency requirement blocks the last default rule $d_i$,
as $B(x) \supset S_1(x) \vee \ldots \vee S_n(x)$ together with
$\neg S_j(a)$ for all $j \neq i$ gives us $S_i(a)$, contradicting the
$\beta$ of $d_i$.
From $\%(S_i(x), B(x), \epsilon_i, \delta_i)$, we know the proportion
of models in which $S_i(a)$ is true, and thus the proportion of models
satisfying this extension, given $E$, is at most $\delta_i$, a small ratio.

\subsection{Sequential Thresholding}
The validity criteria for individual default rules can be extended to
extensions resulted from the application of a chain of default rules.
We can think of the task of regulating the compound effect of multiple
default rules
as adjusting the set of relevant models
by taking into account
the default conclusions of all previously applied rules in the chain of
reasoning.

One way to accomplish this is by {\em sequential
thresholding\/}~\cite{Teng97c}.  The applicability condition
of a default rule $\default{\alpha}{\beta_1, \ldots, \beta_m}{\gamma}$
in an extension can be modified to take into account the validity of the rule.
In addition to requiring that $\alpha$ is provable and that none of
$\neg\beta_1, \ldots, \neg\beta_n$ are
provable,
we require that the default rule
be ``above threshold'', that is, the proportion of relevant
models satisfying the consequent $\gamma$ be greater
than a threshold $1-\epsilon^*$.

The set of relevant models shrinks in a stepwise fashion.
We start out with all the models satisfying the background knowledge and
evidence we have.
As default rules are applied sequentially, the consequent of the
applied rule at each step is taken as true in all subsequent steps.
The relevant models at a particular step
are then those that are consistent with the given
facts and all the consequents of the rules applied in the previous steps.
A default rule, even if it is $\delta$-valid with respect to the background
knowledge, would be blocked from
application if it does not satisfy the thresholding criterion.

In~\cite{Teng97c}, the thresholding metric
is based on a simple probability measure of possible worlds.
We can easily extend this metric to employ the same measure as that used for
evaluating the $\delta$-validity of default rules.

\begin{example}
Reconsider Example~\ref{ex:birds}.
Let us take $\epsilon^* \geq \delta^*$.
We start out with the set ${\cal M}$ of all models satisfying $K$ and $E$.
From $\%(S_1(x), B(x), \epsilon_1, \delta_1)$ we know that $d_1$ is
above threshold, and it satisfies the other conditions for applicability.
Therefore we apply $d_1$ and conclude $\neg S_1(a)$.

Now consider $d_2$.
The set ${\cal M}'$ of relevant models
at this point is a subset of ${\cal M}$; it contains only those
models in ${\cal M}$ that satisfy our new conclusion $\neg S_1(a)$ as well.
We have eliminated the models in which
$S_1(a)$ is true.  Since $S_1(x) \supset \neg S_2(x)$,
and $S_1(x) \supset B(x)$,
all the models eliminated satisfy $B(a)$,
and none satisfies $S_2(a)$.
Thus, in ${\cal M}'$, the number of models satisfying $S_2(a)$
is the same as in ${\cal M}$.  However, the number of models
satisfying $B(a)$ is lower in ${\cal M}'$ as a result of
the addition of $\neg S_1(a)$.  This gives rise to a higher proportion
of models satisfying $S_2(a)$ in ${\cal M}'$ ($\delta_2'$)
than in ${\cal M}$ ($\delta_2$).
If $\delta_2' \leq \epsilon^*$, $d_2$ is still above threshold
after the application of $d_1$, and we can apply it to obtain $\neg S_2(a)$.
Otherwise, $d_2$ is below threshold, and we cannot apply it even though
it was above threshold before the application of $d_1$.

After each step of applying a rule, the set of relevant models
shrinks, and the proportion of $S_i(a)$ of any unapplied rule $d_i$
increases.  After a number of steps, all the remaining rules would be
below threshold, and we thus obtain an extension containing only
a portion of the conclusions that would otherwise be present in the
non-thresholding version of the extension.
\end{example}
Note that the size of $\epsilon^*$
determines how much risk is tolerated in an extension.  The higher
the $\epsilon^*$, the more of the rules can be applied
and the longer they can stay above threshold.
Reiter's non-thresholding version corresponds to the case when $\epsilon^*=1$;
that is, every rule whose associated proportion is above 0 is allowed,
and logical consistency alone determines the rule's admissibility.

\section{Concluding Remarks}

We have developed a notion of validity for default inference based on
model proportions.  A rule is $\delta$-valid if the proportion of models
in which the consequent of the rule is satisfied is greater than $1-\delta$
in the relevant models picked out by the background knowledge, the evidence,
and the
applicability conditions of the default rule.
Given a body of background knowledge $K$, we can systematically generate
candidate default rules and determine which ones are $\delta$-valid
based on the statistical facts known in $K$.
Conflicts between default rules stemming from multiple inheritance
are resolved as a consequence of the validation process.
The result is a set of $\delta$-valid default rules which are
``pre-compiled'' for a given background knowledge base, and can be
reused for different evidence sets without change.

This idea of evaluating the validity of a default rule using model
proportions is extended to extensions generated by
a combinations of rules.  The compound effect is
regulated by a sequential thresholding process,
which blocks the rules whose associated model proportions with respect
to the ``current'' (shrinking) set of models fall below
a particular comfort threshold.  This allows us to use a more reasonable
``closure condition'' for extension than the usual maximal logical strength:
we can refrain from applying rules that would make the extension
satisfiable in only a small set of models, even if the consequent of
the rule is logically consistent with the extension.

Grounding the justification of default rules in model proportions provides
a way to validate the rules empirically, and is a first step towards
automating the learning of default rules from (statistical) data.
One might ask why we need the default rules when we can reason with
the statistics directly.  Default rules provide a succinct and more
understandable characterization of the import of the data, as well as
a smooth articulation of the  information that
may exist in the knowledge base.

\section*{Acknowledgement}
This work was supported by the National Science Foundation STS-9906128,
IIS-0082928, and NASA NCC2-1239.

\end{document}